\PassOptionsToPackage{numbers,sort&compress}{natbib}
\documentclass[sigconf,nonacm]{acmart}

\usepackage[utf8]{inputenc}
\usepackage[T1]{fontenc}
\usepackage{microtype}
\usepackage{booktabs}
\usepackage{tabularx}
\usepackage{amsmath}
\usepackage{xcolor}
\usepackage{url}

\setcopyright{none}
\hypersetup{
  hidelinks,
  pdftitle={Four Ledgers, Not One Score: Responsible Communication of LLM-Judge Calibration in Biomedical ML},
  pdfauthor={Sidi Chang; Peiying Zhu}
}
\providecommand{\doi}[1]{\url{https://doi.org/#1}}

\newcommand{\Pledger}{\mathcal{P}}
\newcommand{\Dledger}{\mathcal{D}}
\newcommand{\Cledger}{\mathcal{C}}
\newcommand{\Rledger}{\mathcal{R}}
\newcommand{\Aledger}{\mathcal{A}}
\newcommand{\Hledger}{\mathcal{H}}

\begin{document}
\acmshorttitle{Four Ledgers, Not One Score}
\acmshortauthors{Chang \& Zhu}
\pagestyle{acmstyle}
\thispagestyle{firstpage}

\twocolumn[
\begin{@twocolumnfalse}
\begin{center}
{\LARGE\bfseries Four Ledgers, Not One Score: Responsible Communication of LLM-Judge Calibration in Biomedical ML\par}
\vspace{1.0em}

\begin{tabular}{@{}c@{\hspace{2em}}c@{}}
{\large Sidi Chang\textsuperscript{*\,$\dagger$}} &
{\large Peiying Zhu\textsuperscript{*}} \tabularnewline
{\small schang@blossomai.co} & {\small peiying@blossomai.co} \tabularnewline
{\small Blossom AI} & {\small Blossom AI} \tabularnewline
{\small San Francisco, CA, USA} & {\small San Francisco, CA, USA}
\end{tabular}
\vspace{1.2em}
\end{center}

\noindent\parbox{\textwidth}{
\small
\noindent\textbf{Abstract}\\[3pt]
Synthetic perturbations appear to offer inexpensive calibration data for LLM evaluators in biomedical ML, where expert review is scarce. Yet a planted mutation key is neither a detector output nor automatically human ground truth. We formalize four distinct ledgers: planted perturbations, independent detector outputs, source-linked human dispositions, and human-added discoveries. We then audit the evaluation design, scoring code, read paths, and current human records of a private synthetic Japanese care-handoff workflow. The factory stored 69 planted error cards across 47 targets. Final review covers 22 targets and contains 22 confirmed imported proposals, 9 rejected proposals, and 79 human-added cards; only 3 reviewed targets are double annotated. Passing imported plant keys to a generic detector scorer yields $22/(22+9)=0.710$ and $22/(22+79)=0.218$. A direct audit identity shows that these values are proposal-confirmation yield and submitted-ledger composition, not judge precision and recall, because no independent detector realization was preserved for the audited proposals in the available records. The audit also finds source-name collisions, row shadowing, forced severity, vacuous ratio defaults, and unsupported zero-support field weights. We contribute a provenance-aware claim audit, a storage contract, and a minimum calibration gate for responsibly communicating biomedical ML capability claims. This single-workflow forensic case is an existence proof of a failure mode, not an estimate of its prevalence: existing human work supports an exploratory audit of synthetic proposals, but not LLM-judge operating characteristics, clinical validity, corpus prevalence, or robust inter-annotator agreement.
\par\medskip
\noindent\textbf{Keywords:} LLM-as-a-judge, biomedical machine learning, evaluation provenance, synthetic perturbations, measurement validity
}

\vspace{1.5em}
\end{@twocolumnfalse}
]

\renewcommand{\thefootnote}{\fnsymbol{footnote}}
\footnotetext[1]{Both authors contributed equally to this research.}
\footnotetext[2]{Corresponding author.}
\renewcommand{\thefootnote}{\arabic{footnote}}
\setcounter{footnote}{0}

\section{Introduction}

LLM judges promise scalable evaluation when biomedical outputs resist exact matching and expert review is expensive. Synthetic data seems to make calibration cheap: plant a known omission or fabrication, ask a judge to find defects, and compare the output with the key. This can be valid only if the plant, the judge output, and the human decision remain separate records.

That separation is easy to lose. A list of planted cards may be imported into a table named for judge results, displayed as \texttt{judgeCards}, and consumed by a precision/recall function. Reviewers can confirm, reject, or add cards. The resulting counts have the algebraic shape of true positives, false positives, and false negatives, but their scientific meaning depends on who produced each card and how it was selected. Metric-shaped values can communicate a biomedical capability that was never measured.

We treat this as a provenance and responsible-communication problem. For each item, we distinguish what the synthetic factory intentionally planted ($\Pledger$), what an independent detector actually emitted ($\Dledger$), how humans disposed of source-linked proposals ($\Cledger,\Rledger$), and what humans discovered beyond them ($\Aledger$). Human-centered evaluation research shows that framing, instructions, rater populations, and task constructs affect reported outcomes \cite{schoch-etal-2020-problem,van-der-lee-etal-2019-best,howcroft-etal-2020-twenty}. Model judges likewise show systematic biases \cite{zheng-etal-2023-judging,chen-etal-2024-humans,fu-etal-2023-large}. Our contribution precedes model choice: an evaluator cannot have its precision or recall estimated when its outputs are not observed as an independent ledger.

We use a private Japanese care-handoff workflow as a forensic case study. Synthetic spoken reports are paired with structured six-field notes, and controlled omissions or fabrications are created for stress testing. We inspect stored aggregates, handoff specifications, scoring semantics, and relevant read-path and annotation-interface code. We do not rank a model or certify a clinical system. We ask what quantities the current records identify and how those quantities should be communicated.

Our contributions are: (i) a four-ledger measurement model and claim-audit identity; (ii) a traceable audit showing what current human work does and does not establish; (iii) implementation mechanisms that turn provenance ambiguity into plausible metrics; and (iv) a minimum storage and reporting gate for judge-calibration claims in biomedical ML.

\section{Related work}

Human evaluation guidance calls for explicit constructs, criteria, rater recruitment, agreement, and uncertainty \cite{van-der-lee-etal-2019-best,howcroft-etal-2020-twenty}. Clinical note evaluation particularly needs grounded criteria that surface omission and hallucination beyond lexical similarity \cite{ben-abacha-etal-2023-investigation,savkov-etal-2022-consultation,zhou-etal-2025-feedback}. Static audio metrics may also weakly predict situated preferences \cite{li-etal-2025-mind}.

LLM judges can correlate with human preferences but remain sensitive to order, style, verbosity, and model relationships \cite{zheng-etal-2023-judging,chen-etal-2024-humans}. Factuality reliability varies across task, prompt, and model \cite{fu-etal-2023-large}; surveys therefore treat LLM-as-judge as a measurement system needing meta-evaluation \cite{li-etal-2025-generation}. Synthetic examples can train quality classifiers or provide weak supervision \cite{wang-etal-2025-train,peng-etal-2024-incubating,peng-etal-2024-text}, but a generated label remains a product of a policy. It can be useful without being an unbiased sample of natural errors, an independent model prediction, or final human truth.

\section{Four-ledger measurement model}

For item $i$, let $\Pledger_i$ be intentionally planted perturbations and $\Dledger_i$ cards independently emitted by detector $g$. A disposition ledger $\Hledger_i$ stores tuples $(q,s,o)$ for reviewed card $q$, explicit source $s\in\{P,D\}$, and outcome $o\in\{\mathrm{confirm},\mathrm{reject},\mathrm{uncertain}\}$. Let $\Cledger_i^s$ and $\Rledger_i^s$ be source-specific confirmed and rejected cards, and $\Aledger_i^s$ cards added beyond the displayed source set. Additions are reviewer assertions unless separately adjudicated. Treating $\Aledger_i^D$ as detector misses additionally requires a documented, sufficiently complete miss-search process.

If all detector proposals are adjudicated, detector micro-precision is
\begin{equation}
\widehat{\mathrm{Prec}}_D=\frac{\sum_i|\Cledger_i^D|}{\sum_i(|\Cledger_i^D|+|\Rledger_i^D|)}.
\label{eq:precision}
\end{equation}
Under the stronger complete-search assumption, micro-recall is
\begin{equation}
\widehat{\mathrm{Rec}}_D=\frac{\sum_i|\Cledger_i^D|}{\sum_i(|\Cledger_i^D|+|\Aledger_i^D|)}.
\label{eq:recall}
\end{equation}
When humans instead review plant proposals, the identified quantity is proposal-confirmation yield,
\begin{equation}
\widehat{Y}_P=\frac{\sum_i|\Cledger_i^P|}{\sum_i(|\Cledger_i^P|+|\Rledger_i^P|)}.
\label{eq:yield}
\end{equation}
The human-added share of the submitted final-review ledger is
\begin{equation}
\widehat{S}_A=\frac{\sum_i|\Aledger_i^P|}{\sum_i(|\Cledger_i^P|+|\Aledger_i^P|)}.
\label{eq:share}
\end{equation}
Neither is a detector operating characteristic.

\paragraph{Audit identity (detector non-identifiability).}
Suppose the stored proposal list supplied to Equations~\ref{eq:precision}--\ref{eq:recall} is $\Pledger_i$, source-$P$ human dispositions are treated as source-$D$ dispositions, and $\Dledger_i$ is not separately observed. Then numerical ``precision'' reduces to $\widehat{Y}_P$ and numerical ``recall'' to the confirmed-plant share of the submitted final-review ledger. Neither identifies an operating characteristic of $g$.

\paragraph{Derivation.}
Substitute $\Pledger_i$ for $\Dledger_i$ and $\Hledger_i^P$ for $\Hledger_i^D$. Confirmed and rejected elements partition reviewed plants, so Equation~\ref{eq:precision} becomes Equation~\ref{eq:yield}. Equation~\ref{eq:recall} becomes $\sum_i|\Cledger_i^P|/(\sum_i|\Cledger_i^P|+\sum_i|\Aledger_i^P|)$. No term depends on an independent realization of $g$; detectors with different outputs are observationally equivalent under the stored record. $\square$

A sensitivity-to-plants measure would require both ledgers and a versioned matching rule $M$:
\begin{equation}
\widehat{\mathrm{Sens}}_{D\mid P}=\frac{\sum_i|M(\Dledger_i,\Cledger_i^P)|}{\sum_i|\Cledger_i^P|}.
\end{equation}
It is not computable when $\Dledger$ was never independently preserved.

\begin{table*}[t]
\centering
\small
\caption{Record meaning follows producer and selection mechanism, not variable name.}
\label{tab:ledgers}
\begin{tabularx}{\textwidth}{p{0.14\textwidth}p{0.16\textwidth}X X}
\toprule
Ledger & Producer & Supports & Does not support alone \\
\midrule
Plant $\Pledger$ & synthetic factory & intervention coverage and reviewed plant yield & detector output or natural prevalence \\
Detection $\Dledger$ & pinned judge & detector proposals and support & correctness without adjudication \\
Disposition $\Hledger$ & human reviewer & source-specific confirmation/rejection & recall without miss search \\
Discovery $\Aledger^s$ & human reviewer & submitted concerns beyond source $s$ & prevalence under partial review \\
\bottomrule
\end{tabularx}
\end{table*}

\section{Case study and audit method}

The case converts synthetic Japanese spoken care reports into Focus, Subjective, Objective, Assessment, Intervention, and Plan fields. Factory logic creates faithful targets and controlled error cards. An internal interface supports human review. We triangulated: (1) a frozen read-only database audit; (2) current final aggregate annotations; (3) rubric and handoff contracts; and (4) scorer, read-path, and interface code. A deterministic count sheet reproduces displayed ratios and Wilson intervals but does not independently query or validate the database.

The factory population contains 47 targets with 69 planted cards: 47 labeled fabricated and 22 omitted. The available records preserve no independent detector realization for proposals used in the audited final workflow. We select one current final annotation per target by an explicit rule: prefer approved; otherwise use the latest completed record. This yields 22 reviewed targets, all from one cohort. Other cohorts with plants have no usable final verification under this rule. We also require proposal payloads to contain the expected error-card shape so newer unrelated generic outputs cannot shadow the intended record.

Separate workflows remain separate: 12 candidate-note corrections and 75 spoken-input realism records, each with one independent label per item. Correction concerns target editing; realism concerns audio plausibility and the oral/written boundary. Neither adjudicates the independent detector ledger required by Equations~\ref{eq:precision}--\ref{eq:recall}.

We report counts, ratios, and Wilson 95\% intervals. Intervals are descriptive at the displayed item or card grain: they ignore clustering within target, non-random selection, and annotator uncertainty. With only three overlapping targets, we do not estimate agreement.

\section{What current records establish}

Among 22 final reviewed targets, humans disposed of 31 imported proposals: 22 confirmed and 9 rejected. Proposal-confirmation yield is 71.0\% (Wilson 95\% interval 53.4--83.9\%); rejection is 29.0\% (16.1--46.6\%). These are single-reviewer dispositions conditional on a selected slice. They can diagnose possible mismatch among plants, rubric, and reviewer, but are not population validity or detector precision.

Reviewers added 79 cards, or 3.59 per reviewed target. The submitted final-review ledger contains 101 cards: 22 confirmed imports and 79 additions. Additions comprise 78.2\% (69.2--85.2\%). This shows that reviewers asserted many concerns beyond displayed plants. It is not a detector false-negative rate because the displayed list was not an independent detector output and miss-search completeness was not established.

Coverage is 22/47 factory targets, or 46.8\% (33.3--60.8\%). Unreviewed targets are not negatives. At card grain, 38 of 69 plants lack a disposition. If all 38 would be rejected or all confirmed, overall plant confirmation ranges from 31.9\% to 87.0\%; this is a deterministic missing-review bound, not a confidence interval. Only 3/22 reviewed targets have two final annotations, an overlap of 13.6\% (4.7--33.3\%).

If imported plants are mislabeled as detector predictions, a generic scorer receives $TP=22$, $FP=9$, and $FN=79$, emitting 0.710 and 0.218. Table~\ref{tab:claims} states what these ratios actually identify.

\begin{table*}[t]
\centering
\small
\caption{Claim contract for the observed ratios.}
\label{tab:claims}
\begin{tabularx}{\textwidth}{p{0.14\textwidth}X X}
\toprule
Ratio & Identified quantity & Unsupported inference \\
\midrule
$22/31$ & reviewed proposal confirmation & judge precision \\
$22/101$ & confirmed-plant share & judge recall \\
$79/101$ & human-added share & detector false-negative rate \\
$22/47$ & observed target coverage & population accuracy \\
$3/22$ & duplicate-review coverage & robust agreement \\
\bottomrule
\end{tabularx}
\end{table*}

Factory cards are 68.1\% fabricated and 31.9\% omitted; additions are 93.7\% fabricated and 6.3\% omitted. The 25.6-point difference in fabricated share is descriptive. Sources have different selection mechanisms and cards cluster within targets, so it does not estimate natural prevalence or a significant detector bias.

Separate workflows do not repair the ledger. In 12 correction items, one rater changed at least one field in 8, marked hallucination in 3, and marked a missing high-risk fact in 2. The 75-item realism workflow rates spoken input and includes mixed target cohorts. Pooling either workflow into target correctness or judge calibration would change the construct.

\section{Implementation failures}

The failure is not one arithmetic bug but a chain of locally plausible choices.

\paragraph{Source/name collision.}
The interface and scorer call proposals \texttt{judgeCards}, while observed final-workflow rows contain imported factory keys. Storage location and variable name do not prove producer identity.

\paragraph{Shape-dependent row shadowing.}
For two targets, later generic payloads could shadow older error-card payloads when a reader selected the newest row regardless of schema. Selecting by required shape and source before recency recovers the intended rows. This is a read-path defect, not model or reviewer quality.

\paragraph{Forced severity.}
The human-addition form does not elicit severity and stores every addition as high. Severity is therefore a software default, not a human or clinical judgment.

\paragraph{Vacuous ratios and unexercised fields.}
The scorer returns 1.0 when a precision or recall denominator is zero and assigns a neutral weight of 1.0 to fields with zero support. Both values can be miscommunicated as perfect performance. Presentation should emit \textsc{na}, support counts, and explicit states for no error, not reviewed, parse failure, and unexercised field.

\paragraph{Construct boundaries.}
The rubric permits conversion from dialectal or casual speech to standard written clinical Japanese. Verbatim evidence checks can mislabel valid normalization as fabrication. Realism items may also omit record-only facts from speech while paired notes retain them. Omission scoring requires per-fact oral-requirement labels.

\begin{table*}[t]
\centering
\small
\caption{Implementation mechanisms that can create overstated biomedical capability claims.}
\label{tab:risks}
\begin{tabularx}{\textwidth}{p{0.20\textwidth}X X}
\toprule
Risk & Observed failure mode & Minimum repair \\
\midrule
Source collision & plants named \texttt{judgeCards} & explicit producer and detector-run ID \\
Row shadowing & newer incompatible payload wins & select schema/source before recency \\
Forced severity & additions stored as high & elicit or store \texttt{unelicited} \\
Vacuous ratio & zero denominator returns 1.0 & \textsc{na} plus numerator/denominator \\
Unexercised field & zero support stored as neutral 1.0 & null value and support mask \\
Oral/record mismatch & note-only facts scored against speech & per-fact source requirement \\
\bottomrule
\end{tabularx}
\end{table*}

\section{Calibration gate}

A calibration-ready system needs four immutable records.

\textbf{Plant ledger:} item and content hashes; plant ID; mechanism and code version; intended type, field, and fact; pre- and post-mutation values; policy and seed; and visibility rules. Plants remain hidden during detector inference and blind review.

\textbf{Detection ledger:} detector card IDs; pinned model, prompt, temperature, and code; input hashes; raw structured output and parse status; predicted type, field, evidence, and confidence where meaningful. An empty detector output is an observed outcome, not a missing row.

\textbf{Disposition ledger:} for every detector card, confirm/reject/uncertain, anonymized reviewer role, rubric version, evidence, duration, round, and adjudication lineage. Revisions append instead of overwrite.

\textbf{Discovery ledger:} human-added cards stored independently. Match them to plants and detections only after blind review using a versioned hierarchy: exact fact ID; compatible type, field, and evidence; then adjudicated semantic match. Preserve one-to-many and many-to-one relations.

Judge precision or recall should be communicated only when: (i) a pinned independent detection ledger exists; (ii) detection was blind to plant keys; (iii) scored proposals are dispositioned; (iv) additions follow a documented search protocol; (v) items are frozen and disjoint from prompt calibration; and (vi) matching is versioned. If a condition fails, the metric is \emph{not identifiable}, not zero or one. Each reported score should include producer, selection mechanism, unit, support, matching rule, uncertainty, and an excluded inference.

This gate does not require a large new benchmark before useful work begins. It requires each limited judgment to retain the meaning needed for its intended estimand. A future study can calibrate the protocol on a small disjoint slice, then freeze it, reporting micro-by-card and macro-by-target results, target-cluster bootstrap intervals, and stratification by error type, field, register, scenario family, and oral requirement.

\section{Limitations and ethics}

Plants remain valuable for testing failure modes, training targeted filters, and auditing matching logic. The observed 29.0\% rejection rate is useful diagnostic feedback, but with one rating per proposal and sparse overlap it signals possible construct mismatch rather than proving a bad intervention. Likewise, 79 additions identify where generation, matching, or definitions deserve attention without establishing that every addition would survive adjudication.

The distinction is most important when expert labels are few. With 22 reviewed targets, a renamed column can dominate the empirical story. Four ledgers let one review support multiple future purposes while preserving conditional meaning: plant confirmation measures selected intervention validity; detector precision measures detector outputs; detector recall additionally depends on a search-and-match process capable of finding misses.

This is one private workflow and a forensic audit, not a prospective benchmark. Its evidence supports an existence-proof claim---that this provenance failure can produce metric-shaped values---rather than an estimate of how often the failure occurs. Reviewed targets are not asserted to be random. Card-level intervals ignore clustering and reviewer uncertainty. Only three targets overlap, additions may not exhaust errors, and historical interface states cannot be replayed. Type mixtures do not estimate natural prevalence. The proposed schema should be tested outside Japanese care handoffs.

The underlying content is wholly synthetic and contains no patient recordings or health-record data. Only de-identified aggregate statistics from an existing lawful internal quality-review workflow are reported; reviewer identities and participant outcomes are not analyzed. Synthetic clinical text can still encode implausible care patterns and stereotypes. Terms such as fabricated and omitted refer to a declared research construct, not patient harm. Audio, transcripts, targets, fact checklists, metadata, and provenance may be offered under controlled academic or commercial access, subject to data-use, security, and applicable third-party terms. Any access package should preserve the four ledgers and must not advertise judge calibration until the gate is met.

\paragraph{Reproducibility artifact.}
The public artifact at \url{https://github.com/pyingzhu/rcmlr-four-ledger-artifact-2026} contains the frozen count sheet, a standard-library script that reproduces all displayed ratios, Wilson intervals, source-conditioned shares, missing-review bounds, and implementation constants, an executed audit notebook, a claim crosswalk, and a machine-readable four-ledger schema with a wholly synthetic fixture. It contains no row content, reviewer identity, private locator, or detector realization. It reproduces the paper's arithmetic but cannot replay the private extraction or repair the missing detection ledger.

\section{Conclusion}

Synthetic errors can make biomedical evaluation cheaper and sharper only if the measurement system remembers who produced each card. Current review confirms 22 of 31 imported proposals and adds 79 further cards, but the resulting 0.710 and 0.218 do not measure an LLM judge because no independent detector realization was preserved for the audited proposals. Four ledgers restore the distinction among intervention design, detector behavior, human disposition, and human discovery. The discipline yields a modest, defensible claim today and a clear path to calibrated evaluation tomorrow.

\bibliographystyle{plain}
\bibliography{references}
\end{document}